\pdfoutput=1
\documentclass[11pt, a4paper, logo]{gdm/googledeepmind}


\pdftrailerid{redacted}

\makeatletter
\renewcommand\bibentry[1]{\nocite{#1}{\frenchspacing\@nameuse{BR@r@#1\@extra@b@citeb}}}
\makeatother

\usepackage{marvosym}
\usepackage[authoryear, sort&compress, round]{natbib}

\usepackage{times}
\usepackage[T1]{fontenc}
\usepackage[utf8]{inputenc}
\usepackage{microtype}
\usepackage{inconsolata}
\usepackage{xcolor}
\usepackage[most]{tcolorbox}
\tcbuselibrary{breakable}
\usepackage{amsmath,amssymb,amsfonts,amsthm}
\usepackage{bm}
\usepackage{booktabs}
\usepackage{multirow}
\usepackage{graphicx}
\usepackage{subcaption}
\usepackage{algorithm}
\usepackage{algpseudocode}
\usepackage{enumitem}
\usepackage{makecell}
\usepackage{xspace}
\usepackage{url}
\usepackage[colorlinks=true,allcolors=blue]{hyperref}  
\hypersetup{
    citecolor=[HTML]{2420cf},
    linkcolor=[HTML]{2420cf}
}

\newtheorem{finding}{Finding}
\newcommand{\method}{\texttt{ReasonAlloc}\xspace}
\newcommand{\rkv}{R-KV\xspace}

\newcommand{\snapkv}{SnapKV\xspace}
\newcommand{\headnum}{H}
\newcommand{\layernum}{L}

\makeatletter
\renewcommand{\Affilfont}{\mdseries\centering} % 单位字体正常且居中
\makeatother
\title{\centering
    ReasonAlloc: Hierarchical Decoding-Time KV Cache Budget Allocation for Reasoning Models
}

\author[1, $\dagger$]{Wenhao Liu}
\author[1, $\dagger$]{Hao Shi}
\author[2, $\dagger$]{Yunhe Li}
\author[1]{Weizhi Fei}
\author[3]{Xiangyuan Wang}
\author[2]{Mengzhe Ruan}
\author[4]{Hanxu Hou}
\author[5]{Peisong Wang}
\author[2]{Linqi Song}
\author[2, \Letter]{Shuang Qiu}

\affil[1]{%Department of Mathematical Sciences, 
Tsinghua University}
\affil[2]{City University of Hong Kong}
\affil[3]{Peking University}
\affil[4]{Shenzhen University of Advanced Technology}
\affil[5]{Institute of Automation, Chinese Academy of Sciences}

\makeatletter
\renewcommand{\maketitle}{\bgroup\setlength{\parindent}{0pt}
    \begin{center}
        {
            {\titlefont \@title\par}%
            \vskip8pt
            {\@author\par}
            \vskip20pt%
        }%
    \end{center}
	\egroup
	{%
		{\abscontent}
	}%
	\thispagestyle{firststyle}
}%

\def\blfootnote{\xdef\@thefnmark{}\@footnotetext}
\makeatother

\begin{document}
\begin{abstract}
Long chain-of-thought (CoT) trajectories in large language model (LLM) reasoning cause severe inference bottlenecks due to rapid key-value (KV) cache growth. Current decoding-time compression methods mitigate this issue via token eviction, but typically assume a uniform budget distribution across all layers and heads. In contrast, existing non-uniform budget allocation methods are predominantly designed for the static prompt prefill phase, and they do not capture the stepwise context demands of autoregressive reasoning. To bridge this gap, we propose \method, a training-free framework that recasts decoding-time KV compression as a hierarchical budget allocation problem. \method operates at two complementary levels: an offline layer-wise preallocation strategy captures an architecture-driven demand pattern which we call ``\textit{Reasoning Wave}'', while an online head-wise strategy reallocates resources during decoding to information-rich heads based on real-time utility. 
Evaluations on mathematical reasoning benchmarks (MATH-500, AIME~2024) using DeepSeek-R1-Distill-Llama-8B, DeepSeek-R1-Distill-Qwen-14B, and AceReason-14B show that \method outperforms uniform-budget R-KV, SnapKV, and Pyramid-RKV (a baseline enforcing a static, monotonically decreasing layer budget), with the largest gains at small budgets (128-512 tokens). \method is plug-and-play with existing token-eviction policies and introduces negligible inference-time overhead.
\end{abstract}
% {\centering \maketitle \par}
\maketitle
\blfootnote{
    \setlength{\parindent}{0pt}% 取消这一段的缩进
    \makebox[0.2em][l]{$^\dagger$} ~~These authors contributed equally to this work. \\
    \makebox[1.3em][l]{}% 用空白占位，让 Contact 和上一行的字母 T 完美对齐
    \Letter ~~Corresponding Author.~~Email: \texttt{shuanqiu@cityu.edu.hk}
}

\begin{figure}[!h]
    \centering
    % TODO(anon): the overview figure (pic/overview.png) still contains the old method label "Adaptive RKV" in the rendered image. Re-export it with the new method name before submission to avoid de-anonymization.
    \includegraphics[width=\textwidth]{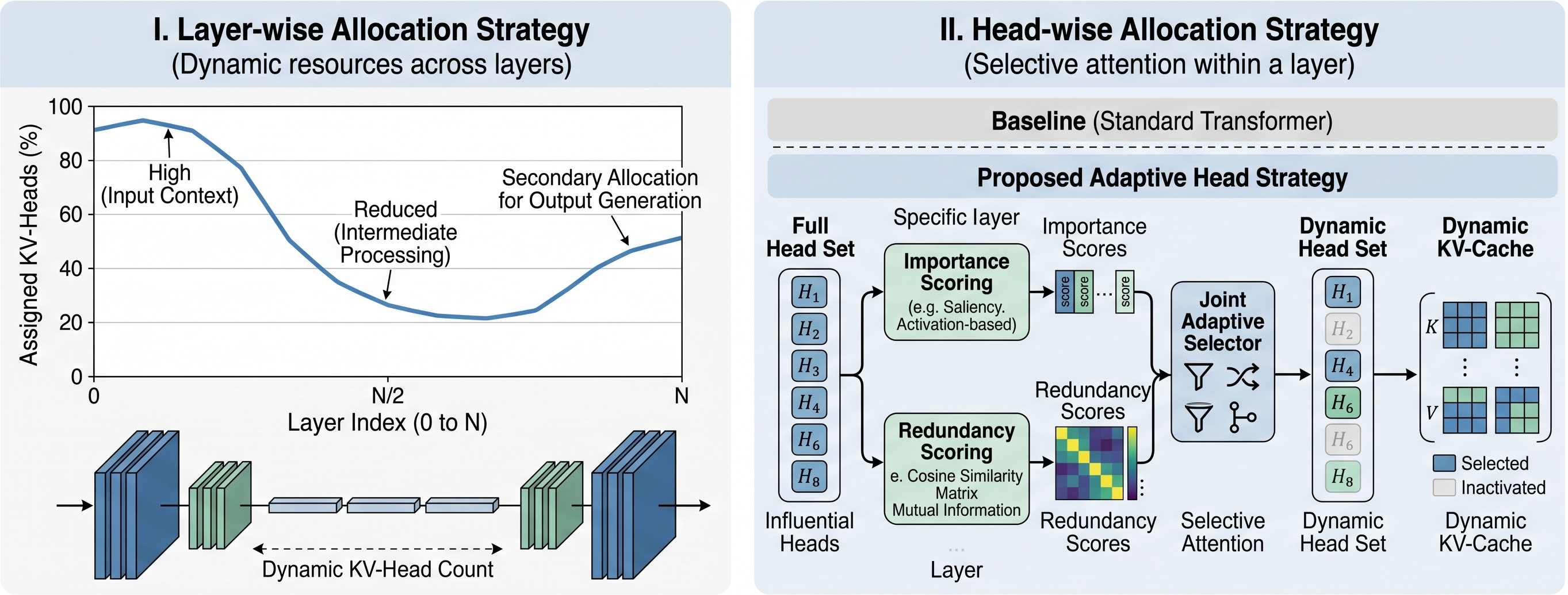}
    \caption{An overview of the proposed \method framework. \textbf{Left (I):} Layer-wise allocation strategy based on offline architecture calibration, demonstrating the non-linear ``Reasoning Wave'' KV demand across layers. \textbf{Right~(II):} Head-wise allocation strategy that dynamically routes KV budgets to distinct attention heads based on real-time importance and redundancy scoring during decoding.}
    \label{fig:overview}
% \vspace{-0.2cm}
\end{figure}

\section{Introduction}

Large reasoning models, particularly those leveraging reinforcement learning (RL) and chain-of-thought (CoT) prompting \citep{wei2022chain, guo2025deepseek}, perform well on complex inference tasks but often generate very long reasoning trajectories. The continuous accumulation of the key-value (KV) cache during decoding becomes a major memory bottleneck and reduces batch size and inference throughput \citep{fatemi2025concise}. Decoding-time KV cache compression has therefore become an active research direction.

Despite recent progress, existing decoding-time compression strategies \citep{rkv, zhang2025lazyeviction} share a common structural assumption: they treat the KV cache budget as \emph{uniformly distributed} across layers and heads. These frameworks focus on \emph{which} tokens to evict, but our empirical analysis (Section~\ref{sec:observations}) shows that the effective KV demand is heterogeneous across layers and heads in reasoning models. To preserve a fraction $\rho$ of the attention mass, some layers and some attention heads require substantially more cache retention than others.

While a separate line of work \citep{cai2024pyramidkv, zhou2024dynamickv} has explored non-uniform budget allocation, these methods are predominantly designed for the prompt prefilling stage. They rely on static mathematical heuristics (e.g., monotonic decay) and do not adapt to the step-by-step utility fluctuations of individual attention heads during autoregressive generation. When these static prefill-centric allocation rules are applied to the decoding phase, they can starve reasoning-critical pathways and lead to accuracy loss, as we show with our Pyramid-RKV baseline.

We therefore frame decoding-time KV compression as a \emph{hierarchical resource allocation} problem: 
%before an underlying policy evicts tokens, the system can choose a per-layer budget and then route that budget across heads based on online decoding-time utility.
before an underlying policy evicts tokens, the system first determines per-layer budgets (\textbf{layer-wise allocation}) and then routes the budgets across heads (\textbf{head-wise allocation}) based on online decoding-time utility.

Motivated by this, we propose \method, a plug-and-play framework that is orthogonal to the underlying token-scoring mechanism and can wrap most decoding-time eviction policies. Our main contributions are:
\begin{itemize}[leftmargin=*,itemsep=2pt,topsep=2pt]
    \item We characterize the heterogeneity of KV cache demand in distilled reasoning LLMs and show that uniform or naively static allocations create a bottleneck for long-context decoding.
    \item We propose \method (Figure~\ref{fig:overview}), a training-free hierarchical allocator that combines offline calibration of a model's approximately task-invariant layer-wise demand (the ``Reasoning Wave'') with an online head-wise router that captures decoding-time utility fluctuations.
    \item On MATH-500 and AIME~2024, using DeepSeek-R1-Distill-Llama-8B and DeepSeek-R1-Distill-Qwen-14B (hereafter R1-Llama-8B and R1-Qwen-14B), along with AceReason-14B, \method surpasses both uniform-budget and fixed non-uniform allocation baselines at the same compression ratios, with the largest gains at small budgets, while adding negligible inference-time overhead.
\end{itemize}

\section{Related Work}
\paragraph{KV Cache Compression.} Existing KV cache optimization methods include token eviction, quantization, token merging, and low-rank compression~\citep{li2024snapkv,zhang2023h2o,liu2024minicache, hooper2024kvquant, chang2024palu, fei2024extending, fei2026efficient}. Among eviction-based methods, \snapkv~\citep{li2024snapkv} ranks tokens using attention-derived importance, while methods like H$_2$O~\citep{zhang2023h2o} and StreamingLLM~\citep{xiao2023efficient} focus on decoding-time retention patterns. \rkv~\citep{rkv} further proposes joint importance-redundancy scoring to improve selection for reasoning-heavy generation. 

\paragraph{Budget Allocation for KV Compression.} A line of work explores non-uniform budget assignment~\citep{cai2024pyramidkv, zhou2024dynamickv}. For instance, PyramidKV~\citep{cai2024pyramidkv} applies a monotonically decreasing arithmetic sequence to the budget motivated by an information funneling hypothesis, while DynamicKV~\citep{zhou2024dynamickv} demonstrates that layer allocation can be highly sensitive to downstream tasks. However, these frameworks are predominantly designed for the prefilling stage and rely on static mathematical heuristics (e.g., arithmetic decay). For reasoning models generating long CoT trajectories, directly applying prefill-oriented static rules to decoding-time eviction frequently starves critical reasoning pathways. For example, simply applying PyramidKV's fixed arithmetic decay to a decoding-time eviction method (which we evaluate later as the \textit{Pyramid-RKV} baseline) leads to performance degradation. In particular, \method bridges this gap by explicitly formulating a hierarchical budget allocation scheme tailored for decoding-time dynamics.

\paragraph{Shortening CoT.} Other works reduce inference costs by shortening CoT via RL, fine-tuning, or prompting~\citep{hou2025thinkprune, yu2025long, chia2023contrastive}. In contrast, \method is a training-free, inference-only framework directly compatible with existing models without altering reasoning behavior.

\section{Preliminaries}\label{sec:pre}
During autoregressive decoding in a Transformer with $\layernum$ layers and $\headnum$ attention heads, the key-value (KV) cache grows linearly, creating a severe memory bottleneck for long reasoning trajectories.

% Consider an autoregressive Transformer with $\layernum$ layers and $\headnum$ attention heads per layer. During decoding, each generated token appends one key and one value vector per layer/head to the KV cache. Let $x_{1:t}$ denote the current sequence after $t$ decoding steps. The memory cost of storing the full KV cache grows linearly with $t$, which becomes prohibitive for long reasoning trajectories.

We focus on decoding-time KV cache compression. At periodic intervals, this method compresses the current cache by retaining only a subset of KV entries for future attention computation. To decide \textit{which} tokens to retain, existing eviction policies typically assign a score to each token. While standard methods (e.g., SnapKV) rely solely on attention-derived importance, reasoning models often generate highly repetitive chain-of-thought texts. To address this, the R-KV framework introduced a joint scoring criterion: a token's utility is determined not only by its historical attention importance but also by its novelty (i.e., penalizing high redundancy compared to other cached tokens). This joint perspective encourages the compressed cache to maintain a diverse and highly informative context.

Let $B$ denote the total allowed token retention budget across the entire model. Let $B^{(\ell)}$ denote the KV budget assigned to layer $\ell$, and let $B^{(\ell,i)}$ denote the budget of head $i$ within layer $\ell$. Standard uniform-allocation methods set $B^{(\ell)} \approx B/L$ and $B^{(\ell,i)} \approx B^{(\ell)}/H$, and then solve a token selection problem under these fixed budgets.

Our goal is to relax this assumption and adapt the budget according to the actual cache demand of each layer and head.

% \section{Observation: Architecture-Driven Heterogeneity in KV Demand}\label{sec:observations}
\section{Architecture-Driven Heterogeneity in KV Demand}\label{sec:observations}

To examine the limitations of uniform budget allocation, we first empirically quantify the true spatial distribution of KV cache demand across the depth of reasoning models. Instead of relying on artificial decaying schedules, we observe the intrinsic retention requirements during the actual generation phase.

\subsection{Profiling Setup and Demand Metric}

We profile three models from the DeepSeek-R1-Distill family \citep{guo2025deepseek}: R1-Llama-8B, R1-Qwen-14B (both introduced above), and additionally DeepSeek-R1-Distill-Qwen-7B (referred to as R1-Qwen-7B). Our evaluation spans datasets with varying complexities: complex mathematical deduction using MATH-500 \citep{math500} and AIME~2024 \citep{aime24}, logical code synthesis via LCB \citep{jain2024livecodebench}, and general long-context understanding via LongBench \citep{bai2025longbench}.

To quantify how much KV cache a specific layer truly needs, we discard simple-average metrics and compute a strict capacity requirement based on the attention distribution. For a given forward pass, we sort all historical tokens in descending order of their attention weights. We then define a layer's \textbf{raw token demand} as the smallest number of top tokens whose cumulative attention sum reaches a target threshold $\rho \in (0, 1)$. Intuitively, this threshold isolates essential reasoning tokens from the long tail of low-utility tokens. We treat $\rho$ as a hyperparameter; its value used in our experiments and a sensitivity analysis are reported in Section~\ref{sec:experiments} and Appendix~\ref{sec:appendix_threshold}. 

To translate this metric into a resource preallocation, we define the \textit{normalized layer-wise KV demand} as a specific layer's raw token demand divided by the sum of raw demands across all layers. This normalization strictly converts absolute token retention counts into a relative percentage, directly reflecting the ideal budget proportion each layer should inherently receive.

% 【新增】跨双栏的 1x3 合并图块
\begin{figure}[t]
    \centering
    \includegraphics[width=\linewidth]{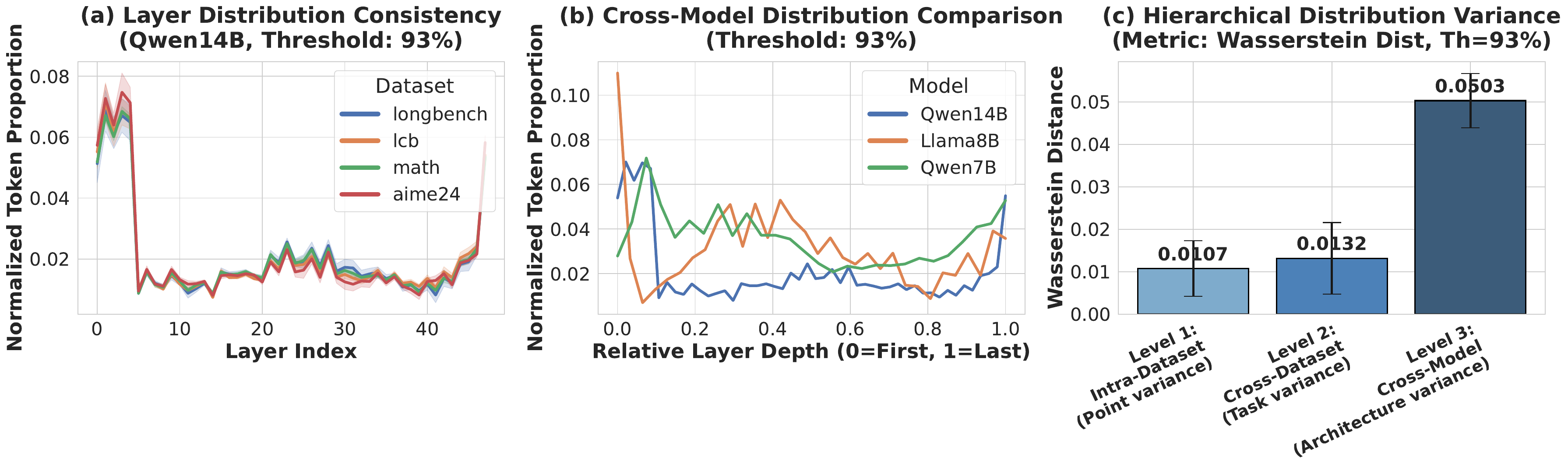}
    \caption{Empirical analysis of layer-wise KV cache demand at $\rho=0.93$ (see Appendix~\ref{sec:appendix_threshold} for the choice of $\rho$). \textbf{(a)} Layer-wise distribution consistency across diverse datasets for Qwen-14B, illustrating the task-invariant ``Reasoning Wave.'' \textbf{(b)} Cross-model architecture divergence, showing distinct demand curves governed by model architectures. \textbf{(c)} Hierarchical distribution variance measured via 1D Wasserstein distance, quantitatively confirming that architectural variance (Level 3) vastly outweighs task or point variance.}
    \label{fig:combined_layer_demand}
\end{figure}

% \subsection{Intra-Model Task Invariance: Reasoning Wave}
\subsection{Intra-Model Task Invariance}
	A common assumption in recent non-uniform compression work is that the layer-wise budget distribution is strongly task-dependent. In our experiments on CoT-distilled reasoning models, however, we find the cross-task variation in layer-wise demand to be small relative to cross-architecture variation (Figure~\ref{fig:combined_layer_demand}c).

As illustrated in Figure~\ref{fig:combined_layer_demand}a, the layer-wise demand curves for R1-Qwen-14B across mathematical reasoning, coding, and general long-context QA overlap closely, with small standard deviations across samples. This is consistent with the CoT distillation process converging these models toward a relatively standardized thinking template, leading to an important empirical observation.
\begin{finding}\label{find:1} For the distilled reasoning models we evaluate, the layer-wise budget proportion is approximately task-invariant and primarily determined by the architecture.
\end{finding}

Furthermore, the required budget does \textit{not} monotonically decrease (as assumed by pyramidal funneling theories). Instead, it manifests as a non-linear \textbf{``Reasoning Wave''}:
\begin{enumerate}[leftmargin=*,itemsep=0pt,topsep=2pt]
    \item \textbf{Shallow Layers (High Demand):} Substantial budget required for broad-spectrum semantic perception and anchoring global context.
    \item \textbf{Middle Layers (Low \& Oscillating Demand):} Demand drops as specific ``reasoning heads'' execute localized logical deductions.
    \item \textbf{Deep Layers (Spiking Demand):} A sudden budget surge right before output generation, reflecting a holistic verification mechanism over the generated reasoning trace.
\end{enumerate}

%\subsection{Inter-Model Divergence \& Quantification}
\subsection{Inter-Model Divergence}
While highly stable within a single model, the demand curves diverge substantially across different architectures (Figure~\ref{fig:combined_layer_demand}b). For instance, R1-Llama-8B exhibits an overwhelming initial spike, whereas R1-Qwen-7B shows a gradual decline with heavier reliance on mid-to-late layers. 

We quantify this via 1-D Wasserstein distance in Figure~\ref{fig:combined_layer_demand}c. We note that the cross-task (Level~2) variance is $0.0132$, while the cross-architecture (Level~3) variance is $0.0503$, roughly $3.8\times$ larger. This supports another critical finding.
\begin{finding} Across the models we evaluate, layer-wise allocation is shaped much more by the model architecture than by the input task, so a single hard-coded arithmetic schedule is unlikely to be optimal across LLM families.
\end{finding}

\subsection{Intra-Layer Head Skew}
Beyond layer-wise variance, extreme heterogeneity exists \textit{within} the layers. Our empirical profiling (detailed in the heatmap provided in Appendix~\ref{sec:appendix_heatmap}) reveals that certain reasoning-critical heads require nearly 500 tokens to reach the $\rho$-attention-mass threshold, while adjacent heads in the exact same layer require fewer than 50. This yields the following observation.
\begin{finding} The per-head KV demand is highly non-uniform within a layer.
\end{finding}
Statically dividing a layer's budget equally among heads wastes memory on redundant heads while starving critical ones. Furthermore, unlike the stable layer-wise preallocation, head utility fluctuates significantly step-by-step during decoding, necessitating an \textit{online} dynamic allocation mechanism.

\section{\method Framework}\label{sec:method}

Based on our empirical findings, we present \method. By calculating budget allocations in parallel with the compression algorithm, \method acts as a plug-and-play upgrade for existing decoding-time token eviction pipelines. It transforms static token-picking into a two-tiered hierarchical resource allocation problem: an offline mapping to the ``Reasoning Wave,'' and an online router to capture decoding-time head fluctuations. A detailed flowchart illustrating the full pipeline is provided in Appendix~\ref{sec:appendix_flowchart}, and Algorithm~\ref{alg:method_framework} summarizes the procedure end-to-end.

\subsection{Generic Token Scoring Base}
To ensure our framework acts as a universally compatible plug-and-play module, we decouple the budget allocation mechanism from the underlying token selection policy. \method operates on an abstracted utility score $S^{(i)}_{j,t}$ assigned to the $j$-th candidate token in the $i$-th attention head at decoding step $t$. This score represents the token's relative retention value during the decoding phase.

Our framework imposes no strict assumptions about the mathematical form of $S^{(i)}_{j,t}$. For instance, it can natively consume with pure attention-derived scores (as used by SnapKV) or temporal frequency metrics (as seen in $\text{H}_{2}\text{O}$). However, because reasoning models generate highly repetitive CoT trajectories, incorporating a redundancy penalty can be beneficial. As a concrete instantiation, which we adopt for our main evaluations, one can define $S^{(i)}_{j,t}$ utilizing the joint importance-redundancy formulation proposed by R-KV~\citep{rkv}:
$
S^{(i)}_{j,t} = \alpha I^{(i)}_{j,t} + (1-\alpha) R^{(i)}_{j,t},
$
where $I^{(i)}_{j,t}$ denotes attention-based importance, $R^{(i)}_{j,t}$ measures novelty, and $\alpha$ is a balancing hyperparameter.

Regardless of the specific scoring formulation adopted, \method utilizes this scoring landscape solely as a signal to dynamically determine the budget allocation $B^{(\ell,i)}$ \textit{before} the underlying algorithm executes the actual token eviction.

\begin{algorithm}[t] 
\small 
\caption{\method Framework}
\label{alg:method_framework}
\setlength{\baselineskip}{1.18\baselineskip}
\begin{algorithmic}[1]
\Require Model $M$, target attention fraction $\rho$, global budget $B$, refresh interval $\Delta$, head-floor coefficient $\mu$

\Statex \emph{\textbf{// Phase 1: Offline Layer-wise Calibration}}
\For{$\ell=1$ \textbf{to} $L$}
    \State Compute $B^{(\ell)}_{\mathrm{raw}}$ at $\rho$ attention mass
\EndFor
\State $\bar{B} \leftarrow B/L$
\State $\{B^{(\ell)}\}_{\ell=1}^{L} \leftarrow \mathcal{R}_L\big(B^{(1)}_{\mathrm{raw}}, \dots, B^{(L)}_{\mathrm{raw}}\big)$

\Statex \emph{\textbf{// Phase 2: Online Decoding with Dynamic Routing}}
\While{generation not finished}
    \State Generate next token at step $t$
    \If{$t \pmod \Delta == 0$}
        \For{$\ell=1$ \textbf{to} $L$}
            \State Pool $S^{(i)}_{j,t}$ over all heads $i$ into $\mathcal{S}^{(\ell)}_{t}$
            \State $\tau \leftarrow \textsc{KthLargest}\big(\mathcal{S}^{(\ell)}_{t},\, B^{(\ell)}\big)$
            \For{$i=1$ \textbf{to} $H$}
                \State $\hat{r}_i \leftarrow \sum_{j \in C_{i,t}} \mathbb{I}(S^{(i)}_{j,t} \ge \tau)$
            \EndFor
            \State $B_{\min}^{(\ell)} \leftarrow \mu \cdot B^{(\ell)}/H$
            \State $\{B^{(\ell,i)}\}_{i=1}^{H} \leftarrow \mathcal{R}_H^{(\ell)}\big(\hat{r}_1, \dots, \hat{r}_H\big)$
            \State Compress cache in layer $\ell$ using $\{B^{(\ell,i)}\}_{i=1}^{H}$
        \EndFor
    \EndIf
\EndWhile
\end{algorithmic}
\end{algorithm}

\subsection{Offline Layer-wise Preallocation}\label{subsec:layer_alloc}
Supported by Finding~\ref{find:1}, we establish a stable target budget for each layer offline. We run a lightweight probing pass on a small set of random prompts $D$. For each layer $\ell$, we calculate the raw budget $B^{(\ell)}_{\mathrm{raw}}$ required to preserve the target attention mass $\rho$.

In reasoning models, unconstrained capacity demands are typically heavy-tailed. Directly normalizing these raw values to a fixed budget is highly unstable: components with high initial demands tend to monopolize the capacity, pushing others to near-zero and causing a permanent starvation state or representation collapse. To resolve this universally across our hierarchical framework, we define a shared robustification operator $\mathcal{R}(\boldsymbol{x};\, \nu,\, [a, b],\, B_{\mathrm{target}})$ that processes an arbitrary raw demand vector $\boldsymbol{x} = (x_1, \ldots, x_n)$ to output a stable budget allocation vector $\boldsymbol{y} = (y_1, \ldots, y_n)$ under a target total budget $B_{\mathrm{target}}$. This operator systematically unifies our smoothing and boundary protection logic via
\begin{align}
\tilde{x}_k &= x_k^{\nu}, \quad \text{where } \nu \in (0,1), \label{eq:smoothing} \\
\hat{y}_k &= \text{clip}\left( B_{\mathrm{target}} \cdot \frac{\tilde{x}_k}{\sum_m \tilde{x}_m},\, a,\, b \right), \label{eq:clipping} \\
y_k &= \left\lfloor B_{\mathrm{target}} \cdot \frac{\hat{y}_k}{\sum_m \hat{y}_m} \right\rfloor, 
\end{align}
where Eq.~\eqref{eq:smoothing} performs power smoothing to compress the dynamic range and dampen extreme spikes, while Eq.~\eqref{eq:clipping} provides explicit boundary protection to guarantee a minimum working memory and cap maximum consumption per component.

For the layer-wise preallocation phase, we route the raw layer demand vector $\boldsymbol{B}_{\mathrm{raw}} = (B^{(1)}_{\mathrm{raw}}, \ldots, B^{(L)}_{\mathrm{raw}})$ through this operator under the global target budget $B$. For brevity, we denote this layer-tier instance as
\begin{equation*}
    \mathcal{R}_L(\boldsymbol{x}) \;:=\; \mathcal{R}\big(\boldsymbol{x};\, \gamma,\, [0.25\bar{B},\, 2\bar{B}],\, B\big),
\end{equation*}
so the layer budgets are obtained by $\{B^{(\ell)}\}_{\ell=1}^{L} = \mathcal{R}_L(\boldsymbol{B}_{\mathrm{raw}})$. We fix $\gamma = 0.5$ (a square-root contraction), as our preliminary experiments demonstrated that this specific exponent optimally balances the dampening of extreme budget spikes with the preservation of necessary dynamic allocation differences, avoiding the need for task-specific tuning. We clip each layer's allocation to $[0.25\bar{B}, 2\bar{B}]$, where $\bar{B} = B/L$ is the uniform average budget per layer. This strict bounding guarantees that the final layer budgets $\{B^{(\ell)}\}_{\ell=1}^{L}$ strictly sum to $B$ while ensuring no single layer is starved or allowed to disproportionately dominate.

% Moreover, we note that for foundation models that may lack strict cross-task consistency, a fully dynamic online layer prefill fallback is further proposed in Appendix~\ref{appendix:dynamic_layer}.

Moreover, to improve the generality of our framework for foundation models that may lack strict cross-task consistency, we further propose a fully dynamic online layer prefill fallback in Appendix~\ref{appendix:dynamic_layer}.

% \subsection{Offline Layer-wise Prior Allocation}\label{subsec:layer_alloc}
% Supported by Conclusion~1, we establish a stable target budget for each layer offline. We run a lightweight probing pass on a small set of random prompts $D$. For each layer $\ell$, we calculate the raw budget $B^{(\ell)}_{\mathrm{raw}}$ required to preserve the target $\rho{=}93\%$ attention mass. 

% The raw demand vector $(B^{(1)}_{\mathrm{raw}}, \ldots, B^{(L)}_{\mathrm{raw}})$ is then routed through the shared robustification module $\mathcal{R}_{\gamma,[0.25\bar{B},\,2\bar{B}]}$ of Section~\ref{subsec:smoothing}: a square-root smoothing ($\gamma{=}0.5$), clipping to $[0.25\bar{B},\,2\bar{B}]$ where $\bar{B}{=}B/L$, and a final proportional rescaling that strictly yields the layer budgets $\{B^{(\ell)}\}_{\ell=1}^{L}$ summing to $B$. The clipping range prevents any single layer from being starved or from disproportionately dominating the global budget.

% \textit{(Note: For foundation models that may lack strict cross-task consistency, a fully dynamic online layer prefill fallback is detailed in Appendix~\ref{appendix:dynamic_layer}).}

\subsection{Online Head-wise Dynamic Routing}\label{subsec:head_alloc}
The online head-wise dynamic routing is a core distinction of \method. Unlike prefill-centric static algorithms, head allocation is performed \textit{online} and refreshed every $\Delta$ generated tokens to capture decoding-time utility shifts.

At step $t$, the candidate pool $C_{i,t}$ for head $i$ includes both historically retained tokens and newly generated tokens. We calculate $S^{(i)}_{j,t}$ for every head-token pair. The dynamic routing executes via three sequential steps:

\paragraph{Step 1: Global thresholding.} We pool scores across all heads in layer $\ell$ into the multiset $\mathcal{S}^{(\ell)}_{t} = \bigcup_{i=1}^{H} \{ S^{(i)}_{j,t} \mid j \in C_{i,t} \}$, and set $\tau$ to its $B^{(\ell)}$-th largest element so that approximately $B^{(\ell)}$ head-token pairs satisfy $S^{(i)}_{j,t}\ge\tau$. Writing $\textsc{KthLargest}(\mathcal{S}, k)$ for the $k$-th largest entry of a multiset $\mathcal{S}$, the threshold is
$
    \tau \;=\; \textsc{KthLargest}\big(\mathcal{S}^{(\ell)}_{t},\, B^{(\ell)}\big).
$

\paragraph{Step 2: Raw per-head demand.} We obtain an unscaled per-head demand $\hat{r}_{i}$ by counting how many tokens in head $i$'s pool exceed the global layer threshold $\tau$:
\begin{equation*}
\hat{r}_{i} = \sum_{j \in C_{i,t}} \mathbb{I}(S^{(i)}_{j,t} \ge \tau),
\end{equation*}
where $\mathbb{I}(\cdot)$ is the indicator function.

\paragraph{Step 3: Shared robustification.} Without a robust defense mechanism, head allocation easily enters a self-reinforcing failure cycle: if a head experiences a transient drop in utility and receives a near-zero budget, it retains fewer historical tokens, which removes the historical context  needed to score relevant future tokens, permanently driving its subsequent allocations to absolute zero. 

To break this starvation loop, we route the raw head demand vector $\boldsymbol{\hat{r}} = (\hat{r}_1, \ldots, \hat{r}_H)$ through the same robustification operator defined in Section~\ref{subsec:layer_alloc}, under the target layer budget $B^{(\ell)}$. By analogy with $\mathcal{R}_L$, we denote this layer-$\ell$ head-tier instance as
\begin{equation*}
    \mathcal{R}_H^{(\ell)}(\boldsymbol{x}) \;:=\; \mathcal{R}\big(\boldsymbol{x};\, \beta,\, [B_{\min}^{(\ell)},\, B^{(\ell)}],\, B^{(\ell)}\big),
\end{equation*}
instantiated with the empirically validated head-level smoothing exponent $\beta=0.5$ and a protective safety-net floor $B_{\min}^{(\ell)} = \mu (B^{(\ell)}/H)$ (where $\mu=0.25$). The final stable head budgets are then $\{B^{(\ell,i)}\}_{i=1}^{H} = \mathcal{R}_H^{(\ell)}(\boldsymbol{\hat{r}})$ and sum to $B^{(\ell)}$. Intuitively, this dynamically routes more KV capacity to heads that are currently information-rich, while preserving essential context structures across all attention heads to prevent representation collapse.

\section{Experiments}
\label{sec:experiments}

We evaluate \method on mathematical reasoning tasks. Since \method is a budget allocator and not a token scorer, we must instantiate it with a specific underlying token scoring policy. For all following experiments, we instantiate the generic utility score $S^{(i)}_{j,t}$ using the state-of-the-art redundancy-aware formulation from R-KV $S^{(i)}_{j,t} = \alpha I^{(i)}_{j,t} + (1-\alpha) R^{(i)}_{j,t}$.

\begin{figure}[htbp]
    \centering
    \includegraphics[width=\textwidth]{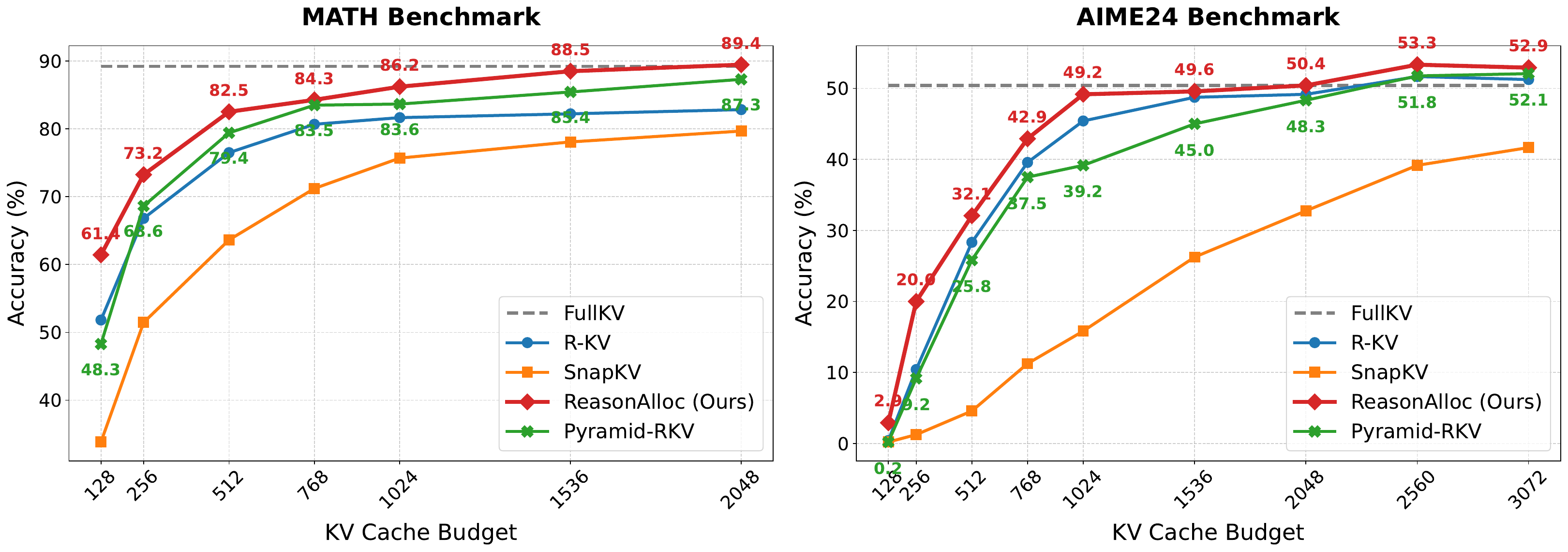}
    \caption{Performance comparison on MATH-500 and AIME~2024 benchmarks across different KV cache budgets (sequence lengths). The proposed \method framework consistently achieves higher accuracy compared to the R-KV baseline and Pyramid-RKV under identical cache constraints, particularly at small budgets.}
    \label{fig:main_benchmark}
\end{figure}

\subsection{Experimental Setup}

\paragraph{Datasets and Models.}
Following common practice for evaluating reasoning-oriented LLMs, we use two mathematical benchmarks: MATH-500 (a curated subset of MATH spanning a range of difficulty levels) and AIME~2024 (competition-level problems). For the underlying models, we evaluate three reasoning-focused architectures known for producing long chain-of-thought trajectories: two distilled variants from the DeepSeek-R1 family (R1-Llama-8B and R1-Qwen-14B), and AceReason-14B.

\paragraph{Baselines.}
We benchmark \method against the following representative KV cache strategies:
\begin{itemize}[leftmargin=*,itemsep=2pt,topsep=2pt]
    \item \textbf{FullKV:} Standard generation without any token eviction, serving as the uncompressed reference.
    \item \textbf{SnapKV:} A static prompt-encoding compression method that selects tokens based on accumulated attention scores, lacking dynamic decoding-time adaptation.
    \item \textbf{R-KV:} The current state-of-the-art decoding-time eviction method. It is our direct baseline because it uses the same joint importance-redundancy scoring as our instantiation, but with \textit{uniform} budget allocation across layers and heads.
    \item \textbf{Pyramid-RKV:} To investigate whether existing prefill-centric allocation heuristics can trivially generalize to decoding, we construct this strong baseline by replacing R-KV's uniform layer allocation with PyramidKV's static, monotonically decreasing arithmetic sequence.
\end{itemize}

\paragraph{Implementation Details.}
To attribute differences to hierarchical allocation rather than to scorer hyperparameters, we keep the token-eviction parameters at the values recommended by R-KV (refresh interval $\Delta = 128$, observation window size $8$, scoring weight $\alpha = 0.1$).
For the unique parameters introduced by \method, we set the attention-mass threshold $\rho{=}0.93$ (selected as justified in Appendix~\ref{sec:appendix_threshold}), the layer power-smoothing exponent $\gamma{=}0.5$, the head power-smoothing exponent $\beta{=}0.5$, the layer clipping range to $[0.25 \bar{B}, 2 \bar{B}]$, and the head floor coefficient to $\mu{=}0.25$ (so $B_{\min}^{(\ell)}{=}0.25\,\bar{B}^{(\ell)}$). The target budgets are allocated through our offline-online hybrid mechanism. We limit the maximum generation length to $16{,}384$ tokens for MATH-500 and $32{,}768$ tokens for AIME~2024.

\paragraph{Evaluation Metrics.}
	We adopt the pass@k evaluation framework~\citep{chen2021evaluating}. Following the recommendations for the DeepSeek-R1-Distill family, we sample with a temperature of $0.6$ and top-$p$ of $0.95$, drawing $k=8$ independent samples per question, and report pass@1 averaged over these $8$ samples. %Each reported accuracy is therefore averaged over $8$ stochastic runs per query rather than a single greedy decoding.

\subsection{Main Results}
The main results for R1-Llama-8B are visualized in Figure~\ref{fig:main_benchmark}, while the complete numerical results for all evaluated models are provided in Appendix~\ref{sec:appendix_main_results}.

\paragraph{Performance under Tight Budgets.}
Across the evaluated budgets, \method achieves higher accuracy than the baselines at most settings, with the largest margins at small budgets. On MATH-500 at a 512-token budget, \method achieves 82.50\% accuracy, compared with 63.62\% for SnapKV and 76.48\% for the uniform R-KV baseline. A similar trend appears on AIME~2024: with a 256-token budget, SnapKV drops to 1.25\%, uniform R-KV reaches 10.42\%, while \method reaches 20.00\%. These results indicate that routing budget toward high-demand layers and heads becomes more important as the total budget tightens.

\paragraph{Limitations of Static Heuristics.}
We additionally evaluate Pyramid-RKV to test whether manually crafted, static non-uniform allocations transfer to long CoT generation. Pyramid-RKV is competitive at moderate budgets but underperforms \method across the evaluated grid (e.g., 79.40\% vs.\ 82.50\% on MATH-500 at budget 512; 39.17\% vs.\ 49.17\% on AIME~2024 at budget 1024). This is consistent with our hypothesis that reasoning models exhibit an architecture-specific ``Reasoning Wave'' rather than a monotonic funnel, and that head utility shifts substantially during generation, so prefill-centric static heuristics do not capture decoding-time variability.

\begin{table}[htbp]
\centering
\small % 缩小表格字号以节省微小的垂直空间
\resizebox{\textwidth}{!}{
\begin{tabular}{l | c c c c c c c c c}
\toprule
\textbf{Ablation Configuration} & \textbf{128} & \textbf{256} & \textbf{512} & \textbf{768} & \textbf{1024} & \textbf{1536} & \textbf{2048} & \textbf{2560} & \textbf{3072} \\
\midrule
R-KV (Baseline, Uniform Alloc.) & 0.42 & 10.42 & 28.33 & 39.58 & 45.42 & 48.75 & 49.17 & 51.67 & 51.25 \\
+ Layer-wise Allocation Only & 0.82 & 13.32 & 30.83 & 38.75 & 42.50 & 43.34 & 48.34 & \textbf{55.00} & 44.18 \\
+ Head-wise Allocation Only & 0.82 & 17.50 & 31.66 & 42.09 & 45.82 & 48.75 & 46.65 & 52.50 & 51.69 \\
\textbf{\method (Full)} & \textbf{2.92} & \textbf{20.00} & \textbf{32.08} & \textbf{42.92} & \textbf{49.17} & \textbf{49.58} & \textbf{50.42} & 53.34 & \textbf{52.92} \\
\bottomrule
\end{tabular}
}
\caption{Ablation study on the AIME~2024 benchmark. We decouple \method to verify the individual contributions of the layer-wise preallocation and the online head-wise dynamic routing.}
\label{tab:ablation}
\end{table}

\section{Efficiency}
\label{sec:efficiency}

All efficiency measurements are run on a single NVIDIA RTX 4090 GPU with 24GB VRAM. KV cache memory is a major bottleneck when serving reasoning models, since it limits both batch size and throughput at long generation lengths. We report system-level throughput across budgets in Figure~\ref{fig:efficiency_throughput}; detailed numbers, including the maximum sustainable batch size, are in Appendix~\ref{sec:appendix_efficiency}.

\begin{figure}[htbp]
    \centering
    % 请将文件名替换为你即将画出的吞吐量/加速比对比图
    \includegraphics[width=\linewidth]{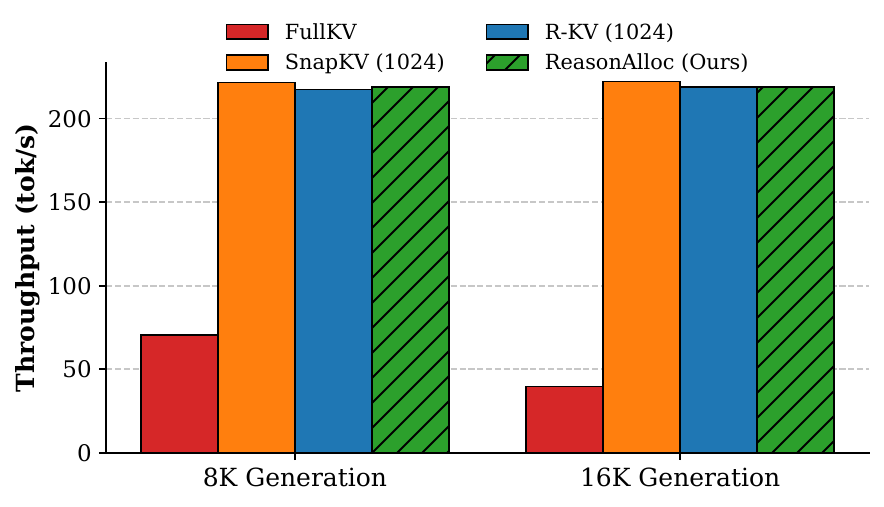} 
    \caption{System-level throughput (tokens/s) for DeepSeek-R1-Distill-Llama-8B. By bounding the KV cache size, \method achieves up to a $5.52\times$ speedup over FullKV at 16K generation lengths while matching the throughput of uniform-budget baselines.}
    \label{fig:efficiency_throughput}
    \vspace{-3mm} % 视情况使用，用于稍微压缩图片下方的空白
\end{figure}

\paragraph{Throughput and Batch Size.}
FullKV memory grows with sequence length, so the maximum batch size drops from 4 at 8K generation to 2 at 16K generation, yielding only 39.66 tokens/s. With a fixed KV cache budget, \method sustains a batch size of 8 in both settings. At a 1024 budget under 16K generation, \method reaches 218.82 tokens/s, a $5.52\times$ speedup over FullKV.

\paragraph{Overhead of Hierarchical Allocation.}
The hierarchical allocation adds negligible overhead. Head-wise routing is invoked only every $\Delta = 128$ decoding steps, and its operations are vectorized over heads. At the 8K/1024 setting, \method runs at 218.78 tokens/s, matching uniform R-KV (217.35 tokens/s). The same parity holds across all evaluated budgets, so the hierarchical allocation does not trade off inference speed for the accuracy gains reported in Section~\ref{sec:experiments}.

% \paragraph{Comparable Inference Throughput.} 
% A crucial finding is the algorithmic efficiency of \method relative to existing uniform compression baselines. Intuitively, the hierarchical allocation process—which involves calculating online global thresholds, evaluating dynamic ratios, and applying power smoothing—might be expected to introduce a computational penalty. However, empirical results demonstrate that the overhead of \method is negligible. 

% For instance, at the 8K/1024 budget setting, \method achieves 218.78 tokens per second, performing completely on par with the 217.35 tokens per second of the uniform R-KV baseline. This throughput parity holds across all evaluated budget constraints. The mathematical operations required for dynamic routing are highly vectorized and do not bottleneck generation, successfully upgrading static token eviction into a cognitive-aware allocator without sacrificing inference speed.

\section{Ablation Studies}
To isolate the contributions of \method's offline layer-wise preallocation and online head-wise dynamic routing, we perform an ablation study on the AIME~2024 dataset (see Table~\ref{tab:ablation}, with performance curves in Appendix~\ref{sec:appendix_ablation}). We compare four configurations: R-KV (Baseline) uses strictly uniform budgets; + Layer-wise Allocation Only applies architecture-calibrated layer budgets but uniform head budgets; + Head-wise Allocation Only retains uniform layer budgets but dynamically routes among heads; and \method (Full) is our complete hierarchical framework.

\paragraph{Component Analysis.} Dynamic head routing alone gives the largest gain at small budgets ($10.42 \to 17.50$ at budget 256), suggesting that protecting the budget of high-utility heads is the dominant factor when the cache is tight. The layer-wise preallocation alone is weaker at mid-range budgets but reaches 55.00 at budget 2560, indicating that the ``Reasoning Wave'' captures a useful global pattern, yet applying it uniformly across heads still leaves within-layer imbalance. \method~(Full) combines both: the layer preallocation sets a stable per-layer budget and online routing resolves the remaining within-layer imbalance, giving the most consistent accuracy across budgets. 

% The full \method method seamlessly synthesizes the strengths of both modules. By utilizing the layer-wise prior to stabilize the global context distribution, and the online head-wise routing to resolve local bottlenecks on the fly, the full framework delivers the most stable and consistently high performance across the entire budget spectrum.

\section{Conclusion}

We presented \method, a hierarchical KV cache allocation framework that formulates decoding-time compression as a resource-routing problem. An offline layer preallocation captures the architectural ``Reasoning Wave,'' while an online head-wise router tracks decoding-time utility, preserving CoT signals under tight memory budgets. As a training-free, plug-and-play layer above existing token-scoring policies, \method offers a practical path toward serving long-context reasoning models more economically.

\bibliographystyle{abbrvnat}
%\nobibliography*
\bibliography{main}

\clearpage
\appendix

\section{Flowchart of \method}
\label{sec:appendix_flowchart}

\begin{figure}[htbp] % 建议加上 htbp 控制浮动位置
    \centering
    \includegraphics[width=\linewidth]{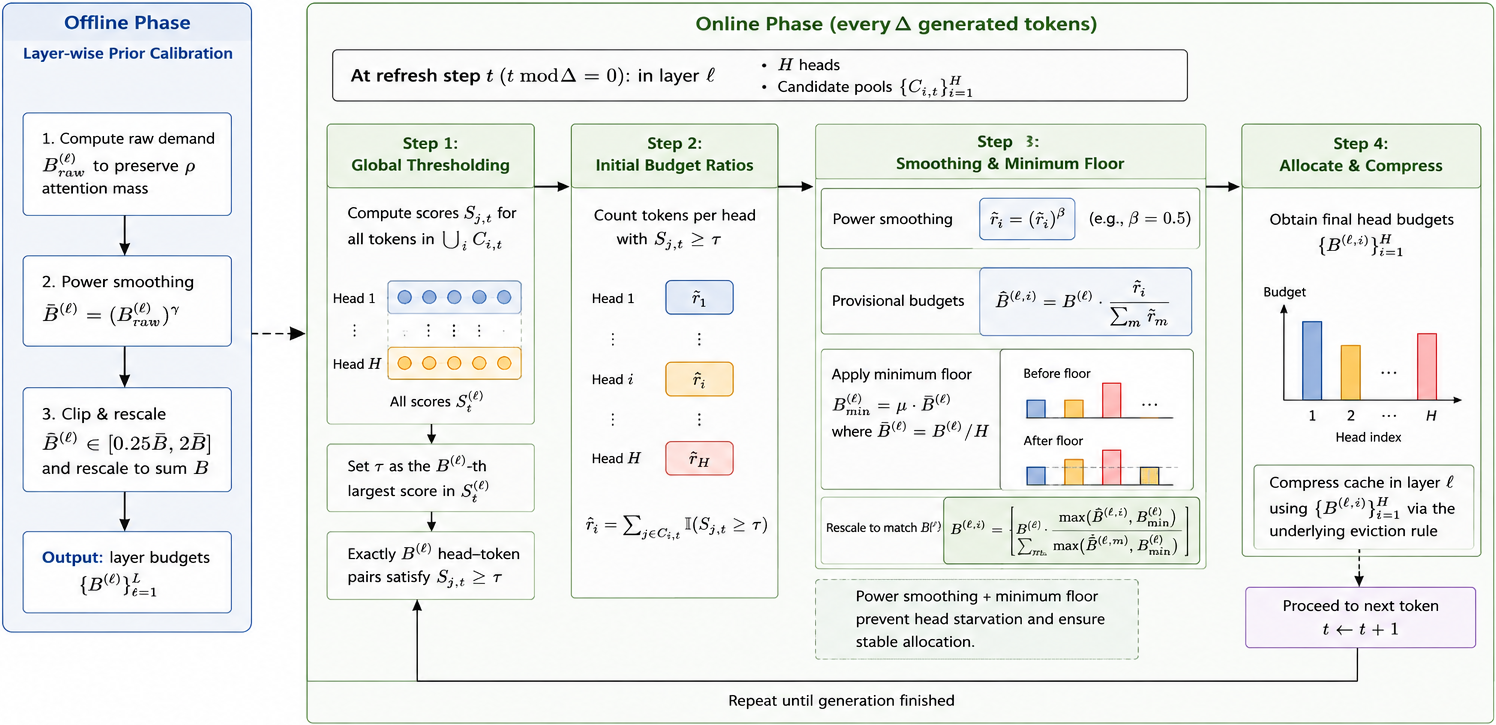}
    \caption{The \method pipeline. \textbf{Left (Offline):} A lightweight calibration pass extracts each layer's raw KV demand to preserve the target attention mass $\rho$; the raw vector is then passed through a shared robustification operator (Section~\ref{subsec:layer_alloc}) and rescaled to obtain a stable per-layer budget $\{B^{(\ell)}\}$. \textbf{Right (Online):} Every $\Delta$ decoding steps, scores $S^{(i)}_{j,t}$ are pooled across heads within layer $\ell$, a single global threshold $\tau$ is set to the $B^{(\ell)}$-th largest score, and per-head raw counts are pushed through the same robustification operator, this time with a head-level minimum floor, to yield the final head budgets $\{B^{(\ell,i)}\}$ that drive the underlying eviction policy.}
    \label{fig:framework}
\end{figure}

Figure~\ref{fig:framework} provides a visual summary of the offline calibration and online dynamic routing pipelines detailed in Section~\ref{sec:method}.

\section{Intra-Layer Head Heterogeneity Heatmap}
\label{sec:appendix_heatmap}
\begin{figure}[htbp]
    \centering
    \includegraphics[width=0.8\linewidth]{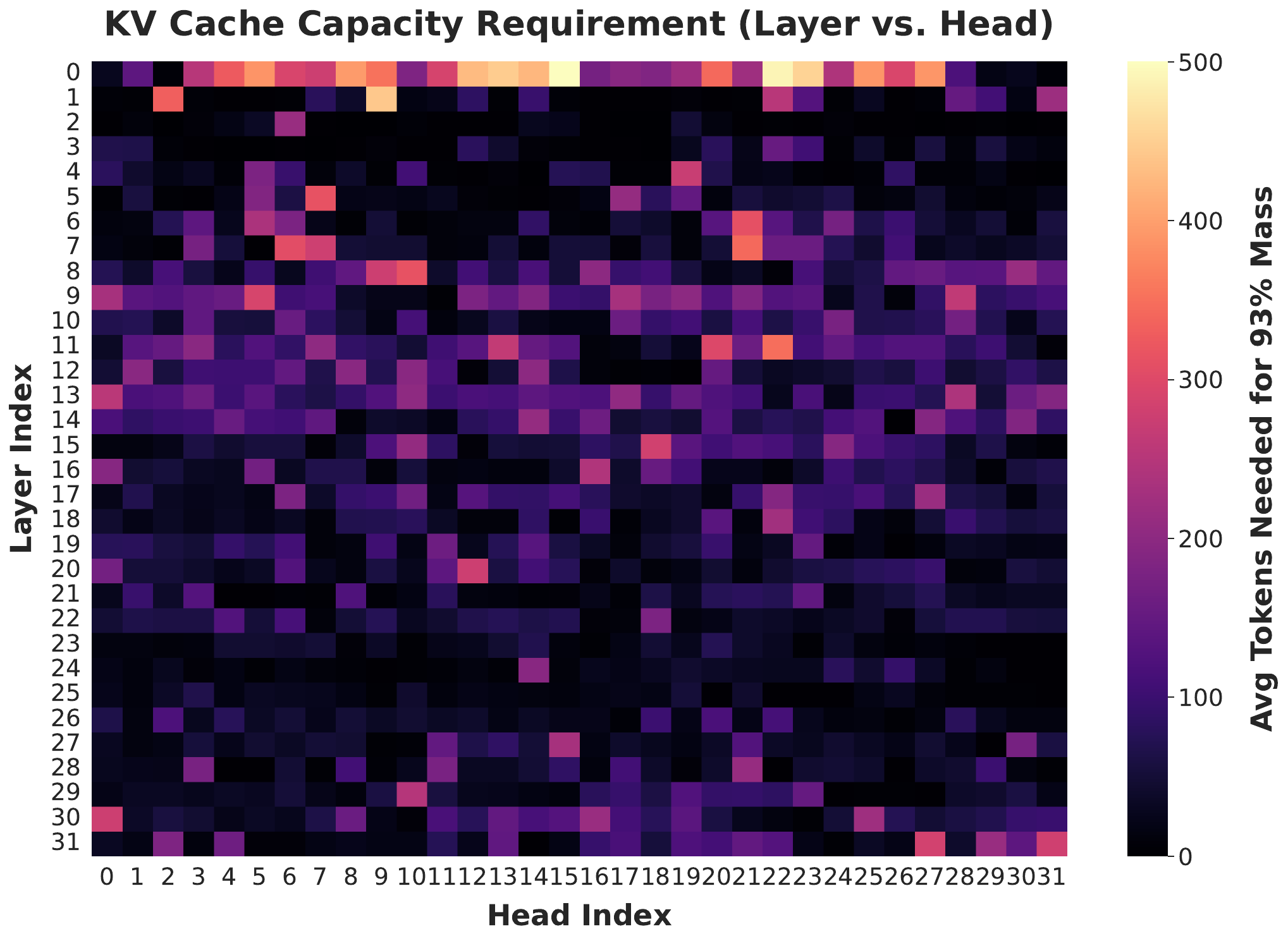}
    \caption{KV Cache Capacity Requirement (Layer vs. Head). The heatmap reveals extreme intra-layer head heterogeneity, with certain reasoning-critical heads requiring nearly 500 tokens to maintain attention mass, while adjacent heads need fewer than 50 tokens.}
    \label{fig:head_heatmap}
\end{figure}

To complement the observations in Section~\ref{sec:observations}, Figure~\ref{fig:head_heatmap} visualizes the extreme intra-layer head heterogeneity.

\section{Fully Dynamic Layer Allocation}
\label{appendix:dynamic_layer}

As discussed in Section~\ref{sec:observations}, while modern CoT-distilled reasoning models exhibit a task-invariant ``Reasoning Wave'' that justifies our offline preallocation, certain general-purpose large language models may have highly task-dependent KV cache demands. For such architectures, a uniform or static offline preallocation may lead to suboptimal token retention. To ensure the universal applicability of \method, we introduce a fully dynamic layer allocation algorithm as a fallback, drawing inspiration from task-aware dynamic allocation methodologies.

\subsection{Task-Aware Online Layer Profiling}
Instead of relying on an offline calibration set, the fully dynamic fallback calculates the layer-wise budget distribution $B^{(\ell)}_{\mathrm{raw}}$ strictly during the prefill phase of the specific inference task. We leverage the attention scores of the most recent tokens (e.g., the current query or the end of the prompt) to estimate the contextual demand of each layer.

Let $ws$ denote the window size of the most recent tokens. For each layer $\ell$, we compute the pooled attention score $A^{(\ell)}$ from the recent window to all preceding tokens in the context. Following the principle of dynamic progressive cache updates, we determine the number of highly attended tokens per layer. Specifically, we apply a global Top-$K$ selection across all layers based on the scores $A^{(\ell)}$ and count the local retention frequency $C^{(\ell)}$ for each layer $\ell$:
\begin{equation*}
    C^{(\ell)} = \sum_{j \in \text{context}} \mathbb{I}\left(A^{(\ell)}_j \ge \tau_{\mathrm{prefill}}\right),
\end{equation*}
where $\tau_{\mathrm{prefill}}$ is the global attention threshold required to retain the target budget during the prefill phase, and $\mathbb{I}(\cdot)$ is the indicator function. 

The raw dynamic budget demand for layer $\ell$ is then proportionally derived from this task-specific retention count:
\begin{equation*}
    B^{(\ell)}_{\mathrm{raw}} = B \times \frac{C^{(\ell)}}{\sum_{m=1}^{L} C^{(m)}}.
\end{equation*}

\subsection{Integration with \method}
Directly allocating the budget based on the raw dynamic count $B^{(\ell)}_{\mathrm{raw}}$ can occasionally be unstable, potentially starving specific layers of their necessary contextual structures due to transient attention spikes. To maintain mathematical and structural consistency with the core \method framework, we route this dynamically derived $B^{(\ell)}_{\mathrm{raw}}$ through the same robustification pipeline introduced in Section~\ref{subsec:layer_alloc}, which we summarize here as four concrete sub-operations:

\begin{enumerate}[leftmargin=*,itemsep=2pt,topsep=2pt]
    \item \textbf{Power Smoothing:} $\tilde{B}^{(\ell)} = (B^{(\ell)}_{\mathrm{raw}})^{\gamma}$ to dampen extreme task-specific spikes and prevent starvation.
    \item \textbf{Initial Scaling:} Globally scale the smoothed budgets to produce intermediate budgets $\hat{B}^{(\ell)}$.
    \item \textbf{Min/Max Clipping:} Clip the intermediate budget $\hat{B}^{(\ell)}$ to the protective range $[0.25\,\bar{B},\,2\,\bar{B}]$ (with $\bar{B}=B/L$) to prevent representation collapse.
    \item \textbf{Secondary Rescaling:} Apply a final proportional rescaling and integer rounding to strictly align with the global budget $B$, yielding the final dynamic layer budget $B^{(\ell)}$.
\end{enumerate}

By replacing the offline probing step with this online task-aware prefill estimation, \method seamlessly transitions into a fully dynamic hierarchical compression algorithm. This fallback ensures robust reasoning preservation even on LLMs with heavily task-dependent attention topographies, successfully accommodating architectural variance with only a marginal computational overhead during the initial prefill phase.

\section{Detailed Experimental Results}
\label{sec:appendix_main_results}
\begin{figure}[H]
    \centering
    \includegraphics[width=\textwidth]{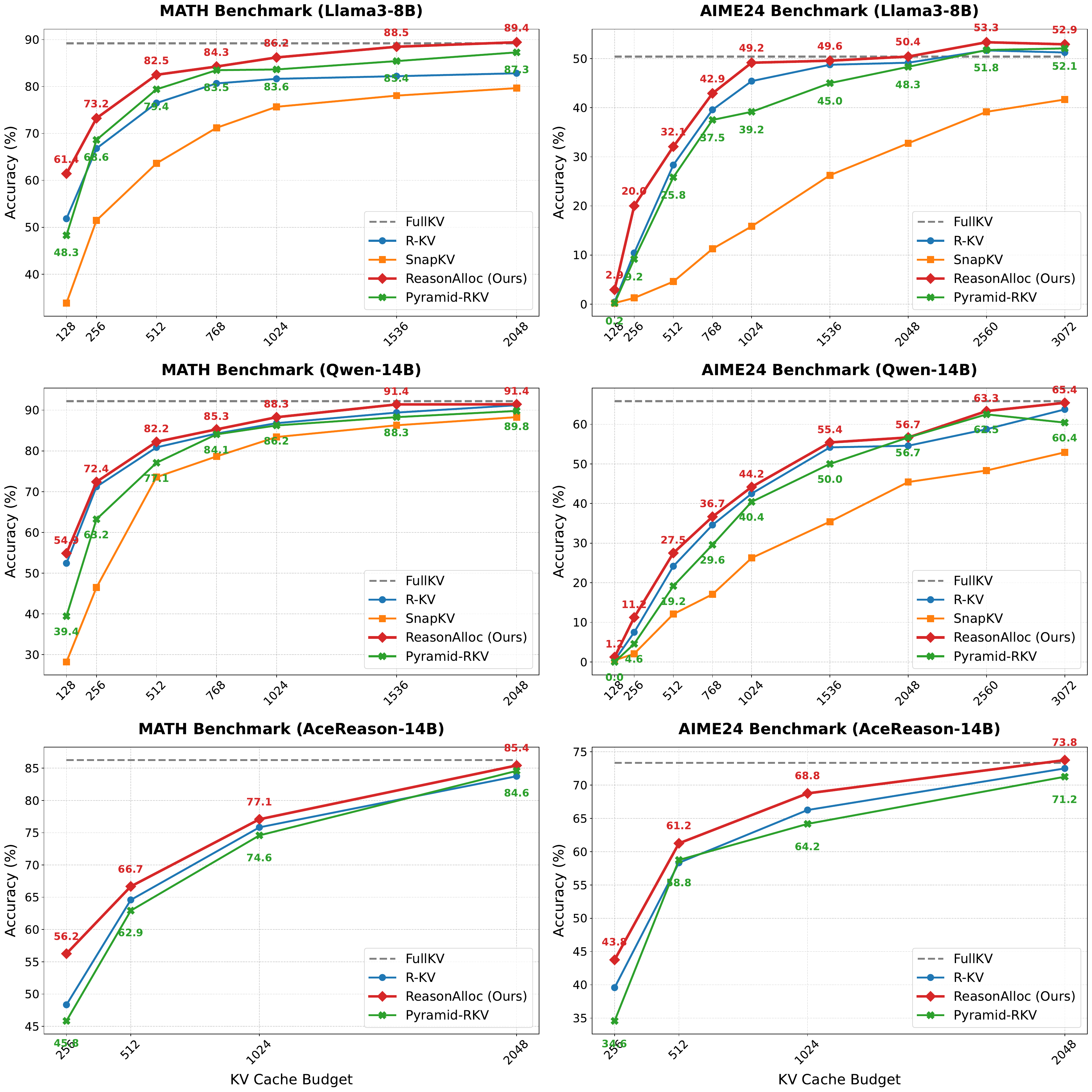}
    \caption{Accuracy on MATH-500 and AIME~2024 for DeepSeek-R1-Distill-Llama-8B, DeepSeek-R1-Distill-Qwen-14B, and AceReason-14B across varying KV cache budgets. Extends the Llama-8B-only view in Figure~\ref{fig:main_benchmark} to all three architectures and both benchmarks. Numerical values are reported in Table~\ref{tab:main_results}.}
    \label{fig:appendix_main_results}
\end{figure}

See Table~\ref{tab:main_results} and Figure~\ref{fig:appendix_main_results}.

\begin{table}[H]
\centering
\resizebox{\textwidth}{!}{
\renewcommand{\arraystretch}{1.12}
\begin{tabular}{l l l | c c c c c c c c c}
\toprule
\textbf{Benchmark} & \textbf{Model} & \textbf{Method} & \textbf{128} & \textbf{256} & \textbf{512} & \textbf{768} & \textbf{1024} & \textbf{1536} & \textbf{2048} & \textbf{2560} & \textbf{3072} \\
\midrule
\multirow{15}{*}{MATH-500} 
& \multirow{5}{*}{R1-Llama-8B}
& FullKV & 89.20 & 89.20 & 89.20 & 89.20 & 89.20 & 89.20 & 89.20 & - & - \\
& & SnapKV & 33.84 & 51.46 & 63.62 & 71.20 & 75.68 & 78.06 & 79.66 & - & - \\
& & R-KV & 51.82 & 66.80 & 76.48 & 80.66 & 81.64 & 82.20 & 82.82 & - & - \\
& & Pyramid-RKV & 48.28 & 68.62 & 79.40 & 83.48 & 83.64 & 85.42 & 87.28 & - & - \\
& & \textbf{\method (Ours)} & \textbf{61.42} & \textbf{73.24} & \textbf{82.50} & \textbf{84.26} & \textbf{86.20} & \textbf{88.48} & \textbf{89.44} & - & - \\
\cmidrule{2-12}
& \multirow{5}{*}{R1-Qwen-14B}
& FullKV & 92.20 & 92.20 & 92.20 & 92.20 & 92.20 & 92.20 & 92.20 & - & - \\
& & SnapKV & 28.20 & 46.48 & 73.56 & 78.64 & 83.40 & 86.28 & 88.28 & - & - \\
& & R-KV & 52.42 & 71.20 & 80.86 & 84.26 & 86.80 & 89.40 & 91.20 & - & - \\
& & Pyramid-RKV & 39.44 & 63.22 & 77.08 & 84.06 & 86.24 & 88.28 & 89.80 & - & - \\
& & \textbf{\method (Ours)} & \textbf{54.86} & \textbf{72.40} & \textbf{82.22} & \textbf{85.28} & \textbf{88.26} & \textbf{91.40} & \textbf{91.44} & - & - \\
\cmidrule{2-12}
& \multirow{5}{*}{AceReason-14B}
& FullKV & - & 86.25 & 86.25 & - & 86.25 & - & 86.25 & - & - \\
& & SnapKV & - & - & - & - & - & - & - & - & - \\
& & R-KV & - & 48.33 & 64.58 & - & 75.83 & - & 83.75 & - & - \\
& & Pyramid-RKV & - & 45.83 & 62.92 & - & 74.58 & - & 84.58 & - & - \\
& & \textbf{\method (Ours)} & - & \textbf{56.25} & \textbf{66.67} & - & \textbf{77.08} & - & \textbf{85.42} & - & - \\
\midrule
\multirow{15}{*}{AIME~2024}
& \multirow{5}{*}{R1-Llama-8B}
& FullKV & 50.42 & 50.42 & 50.42 & 50.42 & 50.42 & 50.42 & 50.42 & 50.42 & 50.42 \\
& & SnapKV & 0.21 & 1.25 & 4.58 & 11.25 & 15.83 & 26.25 & 32.76 & 39.17 & 41.67 \\
& & R-KV & 0.42 & 10.42 & 28.33 & 39.58 & 45.42 & 48.75 & 49.17 & 51.67 & 51.25 \\
& & Pyramid-RKV & 0.21 & 9.17 & 25.83 & 37.51 & 39.17 & 45.01 & 48.33 & 51.76 & 52.08 \\
& & \textbf{\method (Ours)} & \textbf{2.92} & \textbf{20.00} & \textbf{32.08} & \textbf{42.92} & \textbf{49.17} & \textbf{49.58} & \textbf{50.42} & \textbf{53.34} & \textbf{52.92} \\
\cmidrule{2-12}
& \multirow{5}{*}{R1-Qwen-14B}
& FullKV & 65.83 & 65.83 & 65.83 & 65.83 & 65.83 & 65.83 & 65.83 & 65.83 & 65.83 \\
& & SnapKV & 0.42 & 2.08 & 12.08 & 17.08 & 26.25 & 35.42 & 45.42 & 48.33 & 52.92 \\
& & R-KV & 0.42 & 7.50 & 24.17 & 34.58 & 42.50 & 54.17 & 54.58 & 58.75 & 63.75 \\
& & Pyramid-RKV & 0.00 & 4.58 & 19.17 & 29.58 & 40.42 & 49.99 & \textbf{56.67} & 62.50 & 60.42 \\
& & \textbf{\method (Ours)} & \textbf{1.25} & \textbf{11.25} & \textbf{27.50} & \textbf{36.67} & \textbf{44.17} & \textbf{55.42} & \textbf{56.67} & \textbf{63.33} & \textbf{65.42} \\
\cmidrule{2-12}
& \multirow{5}{*}{AceReason-14B}
& FullKV & - & 73.33 & 73.33 & - & 73.33 & - & 73.33 & - & - \\
& & SnapKV & - & - & - & - & - & - & - & - & - \\
& & R-KV & - & 39.58 & 58.33 & - & 66.25 & - & 72.50 & - & - \\
& & Pyramid-RKV & - & 34.58 & 58.75 & - & 64.17 & - & 71.25 & - & - \\
& & \textbf{\method (Ours)} & - & \textbf{43.75} & \textbf{61.25} & - & \textbf{68.75} & - & \textbf{73.75} & - & - \\
\bottomrule
\end{tabular}
}
\caption{Accuracy (\%) of DeepSeek-R1-Distill-Llama-8B, DeepSeek-R1-Distill-Qwen-14B, and AceReason-14B on the MATH-500 and AIME~2024 benchmarks under different KV cache budget constraints. ``-'' denotes configurations that were not evaluated. \method consistently outperforms baseline methods across architectures, particularly at small budgets.}
\label{tab:main_results}
\end{table}

\section{Choice of Attention-Mass Threshold $\rho$}
\label{sec:appendix_threshold}

The attention-mass threshold $\rho$ controls how many top-attended tokens each layer is asked to keep when we estimate its raw demand (Section~\ref{sec:observations}). We do not push $\rho$ to $1.0$ because the attention tail is dominated by near-zero, low-utility weights that inflate every layer's count and wash out cross-layer differences; we also want $\rho$ high enough that the kept tokens still cover the bulk of the attention mass.

\begin{figure}[htbp]
    \centering
    \includegraphics[width=\linewidth]{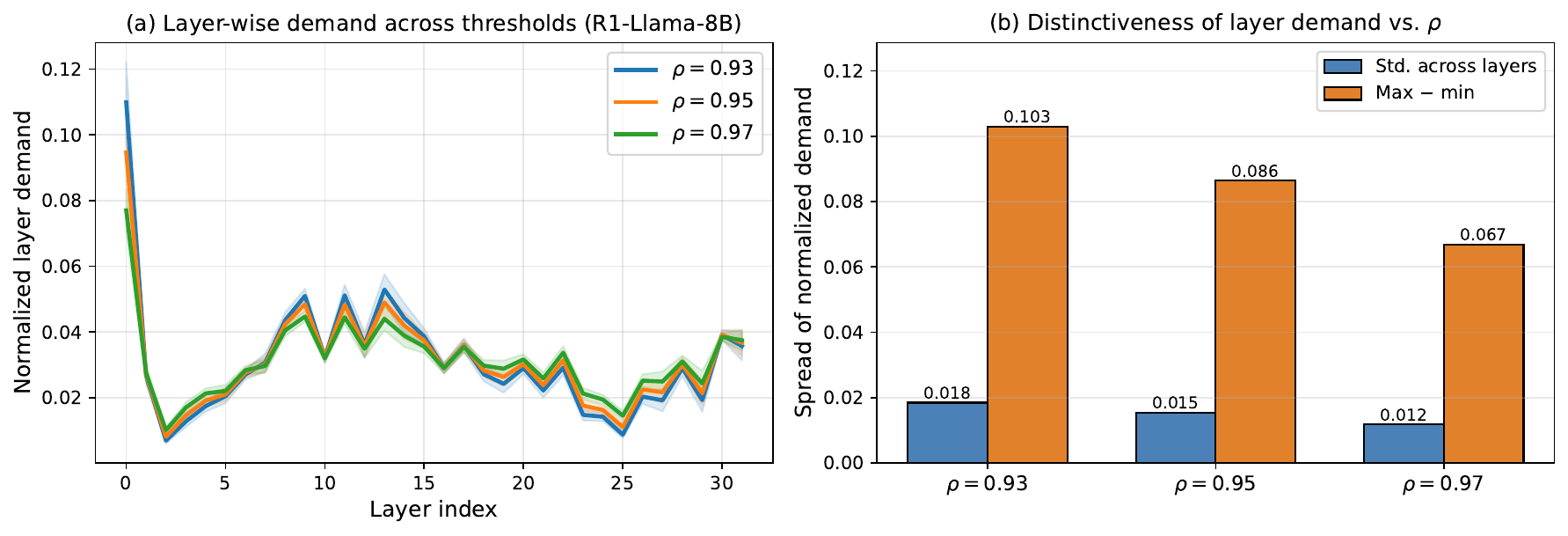}
    \caption{Sensitivity of the layer-wise demand to $\rho$ for R1-Llama-8B. \textbf{(a)} Mean normalized demand per layer at $\rho\in\{0.93,0.95,0.97\}$; shaded bands are $\pm 1$ standard deviation across prompts. The ``Reasoning Wave'' shape is preserved at every threshold. \textbf{(b)} Spread of the mean per-layer demand (lower is flatter / less discriminative); $\rho{=}0.93$ yields the largest spread.}
    \label{fig:threshold_sensitivity}
\end{figure}

Figure~\ref{fig:threshold_sensitivity} compares the normalized layer demand of R1-Llama-8B at $\rho\in\{0.93, 0.95, 0.97\}$, averaged over the same prompts used in Section~\ref{sec:observations}. Two observations support our choice of $\rho{=}0.93$:

\paragraph{(i) Shape is preserved.} All three curves trace the same ``Reasoning Wave'', i.e., the early-layer spike, the mid-layer oscillation, and the deep-layer uptick are present at every threshold (Figure~\ref{fig:threshold_sensitivity}a). The qualitative finding that drives \method, i.e., Finding~\ref{find:1}, therefore does not depend on the exact value of $\rho$.

\paragraph{(ii) $\rho{=}0.93$ is the most discriminative.} As $\rho$ grows, the additional tokens admitted are low-attention tokens that look roughly uniform across layers, so the per-layer counts move toward each other, and the layer-wise contrast flattens. Quantitatively (Figure~\ref{fig:threshold_sensitivity}b), the standard deviation of the mean per-layer demand drops from $0.018$ at $\rho{=}0.93$ to $0.012$ at $\rho{=}0.97$, and the max$-$min spread drops from $0.103$ to $0.067$. A more discriminative signal yields a sharper offline preallocation, which is the actual input \method consumes.

Among the values we examined, $\rho{=}0.93$ both retains a large fraction of the attention mass and preserves the cross-layer contrast that the layer-wise preallocation depends on; the accuracy gains reported in Section~\ref{sec:experiments} are obtained with this choice. We did not perform a full sweep of $\rho$ on downstream accuracy, and we expect nearby values to behave similarly given the shape consistency in (i).

\section{Detailed Ablation Study Results}
\label{sec:appendix_ablation}
\begin{figure}[H]
    \centering
    \includegraphics[width=0.8\linewidth]{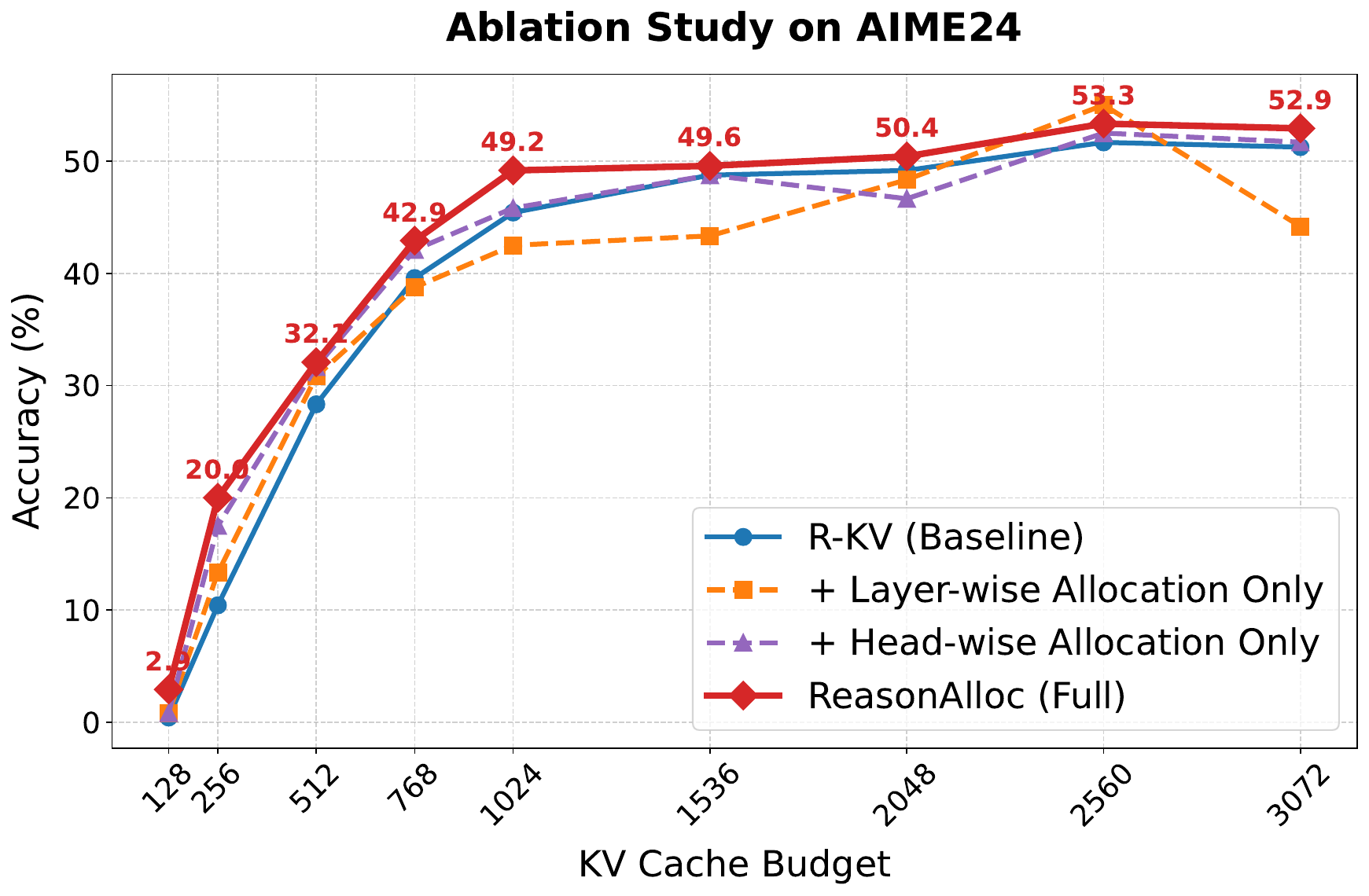}
    \caption{Ablation study of \method on the AIME~2024 dataset. We evaluate the individual contributions of head-wise and layer-wise allocation strategies. The results demonstrate that combining both dimensions (\method Full) yields the optimal performance.}
    \label{fig:ablation_study}
\end{figure}

See Figure~\ref{fig:ablation_study}.

\section{Case Study on Complex Mathematical Reasoning}
\label{sec:appendix_case_study}

To intuitively demonstrate the superiority of \method over the baseline RKV in tasks requiring rigorous, long-context reasoning, we present a comparative case study using Problem 11 from the 2024 AIME II. 

The problem requires finding the number of triples $(a,b,c)$ of nonnegative integers satisfying $a + b + c = 300$ and $a^2b + a^2c + b^2a + b^2c + c^2a + c^2b = 6,000,000$. The correct answer is 601.

The baseline model correctly initiates the algebraic substitution, defining $Q = ab + bc + ca$ and $P = abc$, and arrives at the intermediate state $100Q - P = 2,000,000$. However, as the reasoning context grows, it loses the global mathematical constraint and degrades into heuristic guessing and brute-force trials:

\begin{tcolorbox}[colback=gray!5!white, colframe=gray!50!black, title=\textbf{Baseline RKV Generation Excerpt (Failed)}, breakable, fonttitle=\small, fontupper=\small]
\textit{...} \\
\textit{So, $100Q - P = 2,000,000$. Hmm, that's an interesting equation. Let me write it as: $P = 100Q - 2,000,000$.} \\
\textit{... [Context lost] ...} \\
\textit{Let me divide both sides by 3 to simplify: $a^3 - 300a^2 + 30,000a - 1,000,000 = 0$. Hmm, that's a cubic equation. Maybe I can factor it or find rational roots... Let me try $a = 100$... Not a root. How about $a = 120$... Not a root...} \\
\textit{...} \\
\textit{Case 1: One of the variables is zero... Case 2: All variables are equal... Thus, total solutions: $6 + 1 = 7$. Final Answer: 7.}
\end{tcolorbox}

Our proposed \method framework effectively maintains the logical chain over a long context, successfully bridging the intermediate constraints with the global cubic polynomial to achieve the correct analytical generalization:

\begin{tcolorbox}[colback=gray!5!white, colframe=gray!50!black, title=\textbf{\method Generation Excerpt (Succeeded)}, breakable, fonttitle=\small, fontupper=\small]
\textit{...} \\
\textit{So, we have $P = 100 Q - 2,000,000$.} \\
\textit{We also know from Vieta's formulas that for a cubic equation with roots $a, b, c$: $t^3 - 300t^2 + Qt - P = 0$.} \\
\textit{Substituting $P$ into the cubic equation: $t^3 - 300t^2 + Qt - (100Q - 2,000,000) = 0$.} \\
\textit{...} \\
\textit{So x = 100 is a root... Therefore, the cubic factors as $(x - 100)(x^2 - 200 x + (Q - 20,000)) = 0$.} \\
\textit{... [Successfully derives the general form and permutations] ...} \\
\textit{Therefore, the number of ordered triples is 601. Final Answer: 601.}
\end{tcolorbox}

\subsection{Analysis of Reasoning Dynamics}
The divergence in the reasoning trajectories vividly illustrates the KV cache compression bottleneck. The baseline RKV relies on a uniform budget allocation across all layers and heads. As the generation length increases, the model is forced to evict older tokens to satisfy the rigid budget constraint. Consequently, critical algebraic bridge tokens, such as the polynomial coefficients derived earlier in the prompt, are purged. Deprived of these constraints, the baseline model hallucinates non-existent roots and falls back on naive case-by-case guessing, ultimately yielding an incorrect count of 7.

With \method on the same problem, allocation prioritizes layers and heads whose top tokens have larger mass under the $\rho$-attention threshold. In this trace, tokens encoding the substitution $P = 100Q - 2,000,000$ and the symmetric-polynomial structure are scored highly and are more likely to remain in cache as decoding progresses. In this example, this is sufficient for the model to perform the factorization $(t-100)$ explicitly and reach the correct answer $601$. We note that this is a single qualitative example, rather than a systematic analysis of failure cases.

\section{Detailed System Efficiency Metrics}
\label{sec:appendix_efficiency}

To complement the throughput visualization in Section~\ref{sec:efficiency}, Table~\ref{tab:throughput_comparison} provides the comprehensive system-level efficiency metrics, including maximum sustainable batch sizes and total decoding times across different generation lengths and budgets.

\begin{table}[ht] % 建议用 [ht] 允许页面更灵活地放置表格
\centering
\small
\renewcommand{\arraystretch}{1.12}
\setlength{\tabcolsep}{4pt} % 核心修改 1：将默认的列间距从 6pt 缩小到 4pt
\begin{tabular}{llccccc}
\toprule
% 核心修改 2：利用 makecell 将较宽的表头折叠为两行
\makecell[c]{\textbf{Gen.} \\ \textbf{Length}} & 
\textbf{Method} & 
\textbf{Budget} & 
\makecell[c]{\textbf{Max} \\ \textbf{Batch}} & 
\makecell[c]{\textbf{Throughput} \\ \textbf{(tok/s)}} & 
\textbf{Speedup} & 
\makecell[c]{\textbf{Dec. Time} \\ \textbf{(s)}} \\
\midrule
\multirow{10}{*}{8K} 
& FullKV & -- & 4 & 70.45 & 1.00$\times$ & 465.10 \\
\cmidrule{2-7}
& \multirow{3}{*}{SnapKV} 
& Fixed -- 1024 & 8 & 221.58 & 3.15$\times$ & 295.77 \\
& & Fixed -- 1536 & 8 & 186.41 & 2.65$\times$ & 351.57 \\
& & Fixed -- 2048 & 8 & 163.26 & 2.32$\times$ & 401.41 \\
\cmidrule{2-7}
& \multirow{3}{*}{R-KV} 
& Fixed -- 1024 & 8 & 217.35 & 3.09$\times$ & 301.52 \\
& & Fixed -- 1536 & 8 & 183.54 & 2.61$\times$ & 357.07 \\
& & Fixed -- 2048 & 8 & 161.39 & 2.29$\times$ & 406.07 \\
\cmidrule{2-7}
& \multirow{3}{*}{\textbf{\method}} 
& Dynamic -- 1024 & 8 & 218.78 & 3.11$\times$ & 299.56 \\
& & Dynamic -- 1536 & 8 & 184.62 & 2.62$\times$ & 354.97 \\
& & Dynamic -- 2048 & 8 & 161.91 & 2.30$\times$ & 404.77 \\
\midrule
\multirow{10}{*}{16K} 
& FullKV & -- & 2 & 39.66 & 1.00$\times$ & 826.27 \\
\cmidrule{2-7}
& \multirow{3}{*}{SnapKV} 
& Fixed -- 1024 & 8 & 222.16 & 5.60$\times$ & 589.98 \\
& & Fixed -- 1536 & 8 & 186.47 & 4.70$\times$ & 702.93 \\
& & Fixed -- 2048 & 8 & 162.03 & 4.09$\times$ & 808.95 \\
\cmidrule{2-7}
& \multirow{3}{*}{R-KV} 
& Fixed -- 1024 & 8 & 218.85 & 5.52$\times$ & 598.92 \\
& & Fixed -- 1536 & 8 & 183.81 & 4.63$\times$ & 713.07 \\
& & Fixed -- 2048 & 8 & 160.05 & 4.04$\times$ & 818.93 \\
\cmidrule{2-7}
& \multirow{3}{*}{\textbf{\method}} 
& Dynamic -- 1024 & 8 & 218.82 & 5.52$\times$ & 598.99 \\
& & Dynamic -- 1536 & 8 & 184.33 & 4.65$\times$ & 711.07 \\
& & Dynamic -- 2048 & 8 & 159.70 & 4.03$\times$ & 820.73 \\
\bottomrule
\end{tabular}
\caption{System-level efficiency comparison for DeepSeek-R1-Distill-Llama-8B. By rigorously managing the memory footprint, all compression methods unlock larger batch sizes. Crucially, \method achieves up to a 5.52$\times$ speedup over FullKV at 16K generation lengths, demonstrating that its dynamic allocation introduces negligible computational overhead compared to rigid uniform baselines.}
\label{tab:throughput_comparison}
\end{table}

\section{Reproducibility Details}
\label{sec:appendix_repro}

\paragraph{Models and Licenses.}
We evaluate three publicly released checkpoints from the DeepSeek-R1-Distill family: \texttt{DeepSeek-R1-Distill-Qwen-\allowbreak 7B} ($\sim$7B parameters), \texttt{DeepSeek-R1-Distill-\allowbreak Llama-8B} ($\sim$8B), and \texttt{DeepSeek-R1-Distill-\allowbreak Qwen-14B} ($\sim$14B) \citep{guo2025deepseek}, all distributed under the MIT license. We use the official checkpoints from the model providers' public release without further fine-tuning.

\paragraph{Datasets and Licenses.}
MATH-500 \citep{math500} is a 500-problem evaluation subset derived from the MATH dataset, released for research use. AIME~2024 \citep{aime24} is a public collection of competition problems. LiveCodeBench \citep{jain2024livecodebench} and LongBench v2 \citep{bai2025longbench} are also publicly released for academic research. All datasets are used in their intended evaluation role; no personally identifying information is involved.

\paragraph{Hyperparameters.}
Token-eviction hyperparameters strictly follow the optimal R-KV~\citep{rkv} configuration: refresh interval $\Delta{=}128$, observation window size $8$, importance-redundancy weight $\alpha{=}0.1$. \method-specific hyperparameters: layer power-smoothing exponent $\gamma{=}0.5$, head power-smoothing exponent $\beta{=}0.5$, layer clipping range $[0.25 \bar{B}, 2 \bar{B}]$ with $\bar{B}{=}B/L$, head floor coefficient $\mu{=}0.25$ giving $B_{\min}^{(\ell)}{=}0.25\,\bar{B}^{(\ell)}$ with $\bar{B}^{(\ell)}{=}B^{(\ell)}/H$ (see Section~\ref{sec:method}). These values were selected by manual probing on a single AIME~2024 prompt and set fixed for all experiments.

\paragraph{Decoding.}
We sample with temperature $0.6$ and top-$p$ $0.95$, drawing $k{=}8$ independent samples per query and reporting pass@$1$.
The maximum generation length is 16{,}384 tokens for MATH-500 and 32{,}768 tokens for AIME~2024.

\paragraph{Compute Infrastructure.}
Accuracy experiments (Table~\ref{tab:main_results}, Table~\ref{tab:ablation}) are run on a server with $8\times$ NVIDIA H20 GPUs; throughput and batch-size measurements (Table~\ref{tab:throughput_comparison}, Figure~\ref{fig:efficiency_throughput}) are run on a single NVIDIA RTX 4090 (24~GB VRAM) to reflect a consumer-grade single-GPU deployment scenario. We use PyTorch with the HuggingFace Transformers serving stack; KV cache compression hooks are implemented as a drop-in modification of the attention forward pass. %Total wall-clock for the full evaluation grid is on the order of a few hundred GPU-hours.

\paragraph{Calibration Procedure.}
Offline layer-wise calibration: a single forward pass on $\leq 8$ randomly chosen prompts from a held-out AIME-style set, recording, per layer, the minimum top-$k$ tokens whose cumulative attention sum reaches $\rho{=}0.93$. The resulting per-layer demand vector is passed through the robustification pipeline (Section~\ref{subsec:layer_alloc}) once and cached as a constant for all subsequent inference.

\paragraph{Reporting and Randomness.}
All accuracy values are pass@1 computed using $k{=}8$ samples per query under a single decoding seed. Throughput values are obtained by averaging over $\geq 3$ generation passes. We report the mean value, omitting variance because per-run jitter is below 1\% of the reported tokens/s.

% \section{Parallel Work}
% AURA-KV exploits the sparsity and low-rank structure of long-context KV representations to compress the KV cache during the prefill stage~\citep{anonymous2026aurakv}.

\end{document}